\RequirePackage{fix-cm}

\documentclass[preprint,review,5p,times,twocolumn]{elsarticle}       
\usepackage{lineno,hyperref}
\usepackage[T1]{fontenc}
\usepackage{color}
\usepackage[utf8]{inputenc}
\usepackage{microtype}
\usepackage{newtxmath}
\usepackage[russian,english]{babel}
\usepackage{booktabs} 
\usepackage{graphicx}
\usepackage{url}  
\usepackage{tabularx}
\DeclareMathAlphabet{\mathpzc}{T1}{pzc}{m}{it}
\usepackage{euscript}
\usepackage[toc,page]{appendix}

\journal{Speech Communication}
\biboptions{authoryear}
\begin{document}

\begin{frontmatter}


\title{Text Normalization using Memory Augmented Neural Networks}



\author[mymainaddress]{Subhojeet Pramanik\corref{mycorrespondingauthor}}
\ead{subhojeet.pramanik2015@vit.ac.in}

\author[mymainaddress]{ Aman Hussain }
\cortext[mycorrespondingauthor]{Corresponding author}
\ead{aman.hussain2015@vit.ac.in}

\address[mymainaddress]{VIT University, Vandaloor-Kelambakkam Road, Chennai, Tamil Nadu, India}

\begin{abstract}
We perform text normalization, i.e. the transformation of words from the written to the spoken form, using a memory augmented neural network. With the addition of dynamic memory access and storage mechanism, we present a neural architecture that will serve as a language-agnostic text normalization system while avoiding the kind of unacceptable errors made by the LSTM-based recurrent neural networks. By successfully reducing the frequency of such mistakes, we show that this novel architecture is indeed a better alternative. Our proposed system requires significantly lesser amounts of data, training time and compute resources. Additionally, we perform data up-sampling, circumventing the data sparsity problem in some semiotic classes, to show that sufficient examples in any particular class can improve the performance of our text normalization system. Although a few occurrences of these errors still remain in certain semiotic classes, we demonstrate that memory augmented networks with meta-learning capabilities can open many doors to a superior text normalization system.
\end{abstract}


\end{frontmatter}

\section{Introduction}
The field of natural language processing has seen significant improvements with the application of deep learning. However, there are many unsolved challenges in NLP yet to be solved by the prevalent deep neural networks. One of the simple but interesting challenges lies in designing a flawless text normalization solution for Text to Speech and Automatic Speech Recognition systems. Unlike many other problems being solved by neural networks, the tolerance for unacceptable or "silly" errors in text normalization systems is very low. Prevailing neural architectures in \cite{Sproat2017RNN} produce near perfect overall accuracy on such a problem. But there is a caveat. Whereas deciding whether a word needs to be normalized or not turns out to be an easier problem since it mostly falls back on classifying the semiotic class of the token, the actual challenge lies in generating the normalized form of the token. However, such instances which require normalization are usually very sparse which might result in high overall accuracy. Hence, these specific cases must get the highest attention when evaluating the performance of any text normalization system. Yet the existing models are prone to making "silly" mistakes which are extremely non-trivial and mission-critical to TTS (Text-to-Speech) and ASR (Automatic Speech Recognition) systems. Finite-state filters, which perform simple algorithmic steps on the normalized tokens, are then used to correct such errors and "guide" the model. Although developing such FST grammars is a lot simpler than constructing a fully fledged finite-state text normalization system, they do require some human expertise and domain knowledge of the language involved.

We, therefore, ask if there is a way to circumvent the requirement of human involvement and language expertise, and instead design a system that is language agnostic and learns on its own given enough data. Although the most commonly used neural networks are adept at sequence learning and sensory processing, they are very limited in their ability to represent data structures and learn algorithms on its own. There have been recent advancements in memory augmented neural network architectures such as the Differentiable Neural Computer, with dynamic memory access and storage capacity. Such architectures have shown the ability to learn algorithmic tasks such as traversing a graph or finding relations in a family tree. Normalization of semiotic classes of interest, particularly those containing numbers and measurement units, can be performed using basic algorithmic steps. While neural network architectures such as LSTM work sufficiently well in machine translation tasks, they have shown to suffer in these semiotic classes which require basic step-by-step transduction of the input tokens \citep{Sproat2017RNN}.  There is hope that neural network architectures with memory augmentation will be able to learn the algorithmic steps (meta-learning), similar to the Finite-state filters, but without any human intervention or the external knowledge about the language and its grammar. With such memory augmentation, the network should be able to learn to represent and reason about the sequence of characters in the context of the text normalization task.

We begin by defining the challenges of text normalization to try to understand the reason behind these "silly" mistakes. Then, after a brief overview of the prior work done on this topic, we describe the dataset released by \cite{Sproat2017RNN} which has been used in a way so as to allow for an objective comparative analysis. Subsequently, we delve into the theoretical and implementation details of our proposed system. Thereafter, the results of our experiments are discussed. Finally, we find that memory augmented neural networks are indeed able to do a good job with far lesser amounts of data, time and resources. To make our work reproducible, open to scrutiny and further development, we have open sourced a demonstration of our system implemented using Tensorflow \citep{tensorflow2015-whitepaper} at \url{https://github.com/cognibit/Text-Normalization-Demo}. 
    
\section{Text Normalization}
\label{gen_inst}

Text normalization is the canonicalization of text from one or more possible forms of representation to a 'standard' or 'canonical' form. This transformation is used as a preliminary step in Text-to-Speech (TTS) systems to render the textual data as a standard representation that can be converted into the audio form. In Automatic Speech Recognition (ASR) systems, raw textual data is processed into language models using text normalization techniques.

For example, a native English speaker would read the following sentence \textit{Please forward this mail to 312, Park Street, Kolkata} as \textit{Please forward this mail to three one two Park Street Kolkata}. However, \textit{The new model is priced at \$312} would be read as \textit{The new model is priced at three hundred and twelve dollars}. This clearly demonstrates that contextual information is of particular importance during the conversion of written text to spoken form.

Further, the instances of actual conversion are few and far between swathes of words which do not undergo any transformation at all. On the English dataset used in this paper, 92.5\% of words remain the same. The inherent data sparsity of this problem makes training machine learning models especially difficult. The quality requirements of a TTS system is rather high given the nature of the task at hand. A model will be highly penalized for making "silly" errors such as when transforming measurements from \textit{100 KG} to \textit{Hundred Kilobytes} or when transforming dates from \textit{1/10/2017} to \textit{first of January twenty seventeen}.

\subsection{Prior Work}

One of the earliest work done on the problem of text normalization was used in the MITalk TTS system \citep{Allen1987From}. Further, a unified weighted finite-state transducers based approach was proposed by \cite{607867}. The model serves as the text-analysis module of the multilingual Bell Labs TTS system. An advancement in this field was made by looking at it as a language modeling problem by \cite{Sproat2001NormalizationON}. For ASR systems requiring inverse text normalization, data-driven approaches have been proposed by \cite{Pusateri2017}. The latest development in this problem space has been by \cite{Sproat2017RNN}. Even though a recurrent neural network model trained on the corpus results in very high overall accuracies, it remains prone to making misleading predictions such as predicting completely inaccurate dates or currencies. Such "silly" errors are unacceptable in a TTS system deployed in production. However, a few of these errors were shown to be corrected by a FST (Finite State Transducer) which employs a weak covering grammar to filter and correct the misreadings.      

\subsection{Dataset}

For the purposes of a comparative study and quantitative interpretation, we have used the exact same English and Russian dataset as used in \cite{Sproat2017RNN}. The English dataset consists of 1.1 billion words extracted from Wikipedia and run through Google's TTS system's Kestrel text normalization system to generate the target verbalizations. The dataset is formatted into 'before' or unprocessed tokens and 'after' or normalized tokens. Each token is labeled with its respective semiotic class \footnote{ALL = all cases; PLAIN = ordinary word (<self>); PUNCT = punctuation (sil); TRANS = transliteration; LETTERS = letter sequence; CARDINAL = cardinal number; VERBATIM = verbatim reading of character sequence; ORDINAL = ordinal number; DECIMAL = decimal fraction; ELECTRONIC = electronic address; DIGIT = digit sequence; MONEY = currency amount; FRACTION = non-decimal fraction; TIME = time expression} such as PUNCT for punctuations and PLAIN for ordinary words. The Russian dataset consists of 290 million words from Wikipedia and is formatted likewise. These datasets are available at \url{https://github.com/rwsproat/text-normalization-data}.

Each of these datasets are split into 100 files. The base paper (\cite{Sproat2017RNN}) uses 90 of these files as the training set, 5 files for the validation set and 5 for the testing set. However, our proposed system only uses the first two files of the English dataset (2.2\%) and the first four files of the Russian dataset (4.4\%) for the training set. To keep the results consistent and draw objective conclusions, we have defined the test set to be precisely the same as the one used by the base paper. Hence, the first 100,002 and 100,007 lines are extracted from the 100th file \textit{output-00099-of-00100} of the English and Russian dataset respectively.  

\section{Background: Memory Augmented Neural Networks}

Traditional deep neural networks are great at fuzzy pattern matching, however they do not generalize well on complex data structures such as graphs and trees, and also perform poorly in learning representations over long sequences. To tackle sequential forms of data, Recurrent Neural Networks were proposed which have been known to capture temporal patterns in an input sequence and also known to be Turing complete if wired properly \citep{Siegelmann:1995:CPN:207270.207284}. However, traditional RNNs suffer from what is known as the vanishing gradients problem \citep{VANISHING_GRAD}. A Long Short-Term Memory architecture was proposed in \citep{Hochreiter:1997:LSM:1246443.1246450} capable of learning over long sequences by storing representations of the input data as a cell state vector. LSTM can be trained on variable length input-output sequences by training two separate LSTM’s called the encoder and decoder \citep{NIPS2014_5346}. The Encoder LSTM is trained to map the input sequence to a fixed length vector and Decoder LSTM generates output vectors from the fixed length vector. This kind of sequence to sequence learning approach have been known to outperform traditional DNN models in machine translation and sequence classification tasks. Extra information can be provided to the decoder by using attention mechanisms \citep{DBLP:journals/corr/BahdanauCB14} and \citep{Luong2015Effective}, which allows the decoder to concentrate on the parts of the input that seem relevant at a particular decoding step. Such models are widely used and have helped to achieve state of the art accuracy in machine translation systems.

However, LSTM based sequence-to-sequence models are still not good at representing complex data structures or learning to perform algorithmic tasks. They also require a lot of training data to generalize well to long sequences. An interesting approach by \citeauthor{DBLP:journals/corr/JoulinM15} presents a recurrent architecture with a differentiable stack, able to perform algorithmic tasks such as counting and memorization of input sequences. Similar memory augmented neural network \citep{DBLP:journals/corr/GrefenstetteHSB15} have also shown to benefit in natural language transduction problems by being able to learn the underlying generating algorithms required for the transduction process. Further, a memory augmented neural network architecture called the Neural Turing Machine was introduced by \citeauthor{Graves2014Neural} that uses an external memory matrix with read and write heads. A Controller network that works like an RNN is able read and write information from the memory. The read and write heads use content and location-based attention mechanisms to focus the attention on specific parts of the memory. NTM has also shown promise in meta-learning \citep{Santoro2016Oneshot} showing that memory augmented networks are able to generalize well to even lesser training examples.

An improvement to this architecture was proposed by \citeauthor{Graves2016Hybrid} called the Differentiable Neural Computer having even more memory access mechanisms and dynamic storage capabilities. DNC, when trained in a supervised manner, was able to store representations of input data as “variables” and then read those representations from the memory to answer synthetic questions from the BaBI dataset \citep{Weston2015Towards}. DNC was also able to solve algorithmic tasks such as traversing a graph or inferring from a family tree, showing that it is able to process structured data in a manner that is not possible in traditional neural networks. Dynamic memory access allows DNC to process longer sentences and moreover, extra memory can be added anytime without retraining the whole network.

\subsection{Differentiable Neural Computer}

A basic DNC architecture consists of a controller network, which is usually a recurrent network coupled with an external memory matrix $M \in \mathbb{R}^{N\times W}$. At each timestep $t$, the controller network takes as input a controller input vector $\chi_t=[x_t;r_{t-1}^1,\dots,r_{t-1}^R]$, where $x_t	\in \mathbb{R}^X$ is the input vector for the time-step $t$ and $r_{t-1}^1,\dots,r_{t-1}^R$ is a set of $R$ read vectors from the previous time step and outputs an output vector $v_t$ and interface vector $\varepsilon_t \in \mathbb{R}^{(W\times R)+3W+5R+3}$. The controller network is essentially a recurrent neural network such as the LSTM. The recurrent operation of the controller network can be encapsulated as in Eqn \ref{controller}:

\begin{equation}\label{controller}
(v_t,\varepsilon_t) = \EuScript{N}([\chi_1;\dots;\chi_t;\theta])
\end{equation}

\noindent where $\EuScript{N}$ is a non-linear function, $\theta$ contains the all trainable parameters in the controller network. The read vector $r$ is used to perform read operation at every time step. The read vector $r$ defines a weighted sum over all memory locations for a memory matrix $M$ by applying a read weighting $w^r \in \Delta_N$ over memory $M$. $\Delta_N$ is the non negative orthant of $\mathbb{R}^N$ with the unit simplex as a boundary.

\begin{equation}
r=\sum_{i=1}^{N} {M[i,\cdot ] w^r[i]} 
\end{equation}

\noindent where the ‘·’ denotes all $j= 1,\dots, W$. The interface vector $\varepsilon_t$ is used to parameterize memory interactions for the next time step. A write operation is also performed at each time-step using a write weighting $w^w \in \Delta_N$ which first erases unused information from the memory using an erase vector $e$ and writes relevant information to the memory using the write vector $v_t$. The overall write operation can be formalized as in Eqn \ref{Memory_update}. 

\begin{equation}\label{Memory_update}
M_t = M_{t-1} \circ (E-w_t^w e_t^\top)+w_t^w v_t^\top  
\end{equation}

\noindent where $\circ$ denotes element-wise multiplication and $E$ is an $N\times W$ matrix of ones. The final output of the controller network $y_t\in \mathbb{R}^Y$ is obtained by multiplying the concatenation of the current read vector $r_t$ and output vector $v_t$ with a $RW\times Y$ dimensional weight matrix $W_r$.

\begin{equation}
y_t=v_t+W_r[r_t^1;\dots;r_t^R]
\end{equation}

The system uses a combination of different attention mechanisms to determine where to read and write at every time-step. The attention mechanisms are all parameterized by the interface vector $\varepsilon_t$. The write weighting $w^w$, used to perform the write operation, is defined by a combination of content-based addressing and dynamic memory allocation. The read weighting $w^r$ is defined by a combination of content-based addressing and temporal memory linkage. The entire system is end-to-end differentiable and can be trained through backpropagation. For the purpose of this research, the internal architecture of the Differentiable Neural Computer remains the same as specified in the original paper \citep{Graves2016Hybrid}. The open-source implementation of the DNC architecture used here is available at \url{https://github.com/deepmind/dnc}.

\subsection{Extreme Gradient Boosting}
Boosting is an ensemble machine learning technique which attempts to pool the expertise of several learning models to form a better learner. Adaptive Boosting, or more commonly known as "AdaBoost", was the first successful boosting algorithm invented by \citeauthor{Freund99ashort}. \citep{breiman1998} and \citep{6790642} went on to formulate the boosting algorithm of AdaBoost as a kind of gradient descent with a special loss function. \citep{10.2307/2674028}, \citep{friedman2001} further generalized AdaBoost to gradient boosting in order to handle a variety of loss functions. 

Gradient Boosting has proven to be a practical, powerful and effective machine learning algorithm. Gradient boosting and deep neural networks are the two learning models that are widely recognized and used by the competitive machine learning community at Kaggle and elsewhere. Tree boosting as implemented by XGBoost \citep{DBLP:journals/corr/ChenG16} gives state-of-the-art results on a wide range of problems from different domains. The most defining factor for the success of XGBoost is its scalability. It has been reported to run more than ten times faster than existing solutions on a single node. The reason behind using gradient boosting for classification first and a normalization model later is two-fold. First, we solve the data sparsity problem by having the deep neural network train only on the tokens which need normalization.  Second, we turn to the incredible scalability and efficiency of the XGBoost model to condense huge amounts of data which makes it more amenable to quick training and experimental iterations.

Now, we give a brief overview of the gradient boosting algorithm and establish the intuition necessary to understand the proposed solution for text normalization. Essentially, gradient boosting adds and fits weak learners in a sequential manner to rectify the defects of the existing weak learners. In adaptive boosting, these "defects" are defined by assigning higher penalty weights to the misclassified data points in order to restrain the new learner from making the previous mistakes again. Similarly, in gradient boosting, the "defects" are defined by the error gradients.

The model is initiated with a weak learner $F(x_i)$, which is a decision stump i.e. a shallow decision tree. The subsequent steps keep adding a new learner, $h(x)$, which is trained to predict the error residual of the previous learner. Therefore, it aims to learn a sequence of models which continuously tries to correct the residuals of the earlier model. The sum of predictions is increasingly accurate and the ensemble model increasingly complex.

To elucidate further, we consider a simple regression problem. Initially a regression model $F(x_1)$ is fitted to the original data points:  $(x_1, y_1), (x_2, y_2),\cdots, (x_n, y_n)$. The error in the model prediction $y_i - F(x_i)$ is called the error residual. Now, we fit a new regression model $h(x)$ to data : $(x_1 , y_1 - F(x_1)), (x_2 , y_2 - F(x_2)), ..., (x_n , y_n - F(x_n))$ where $F(x)$ is the earlier model and $h$ is the new model to be added to $F(x)$ such that it corrects the error residuals $y_i - F(x_i)$.

\begin{equation}
F(x_i) := F(x_i) + h(x_i) 
\end{equation}
\begin{equation}
F(x_i) := F(x_i) + y_i - F (x_i) 
\end{equation}
\begin{equation}
F(x_i) := F(x_i) - 1 \frac{\delta J}{\delta F(x_i)}
\end{equation}
This is quite similar to the method of gradient descent which tries to minimize a function by moving in the opposite direction of the gradient:

\begin{equation}
\theta_i := \theta_i - \rho \frac{\delta J}{\delta \theta_i}
\end{equation}where $\rho$ is the learning rate. However from a more applied perspective, the model of choice for implementing the weak learners in the XGBoost library \citep{DBLP:journals/corr/ChenG16} are decision tree ensembles which consist of a set of classification and regression trees. Further implementation details are discussed in the following section.

\section{Proposed Architecture}
We propose a two-step architecture for text normalization.  For a given token $w_i$ which is to be normalized, the token $w_i$ and some context words $w_{i-k}$ to $w_{i+k}$ are first fed as characters into an XGBoost classification model, where $k$ is the number of context words. The XGBoost classification model is trained to predict whether a particular word is to be normalized or not.

For those words which require normalization, a second model is used. We propose a novel sequence-to-sequence architecture based on Differentiable Neural Computer. This second model uses the Encoder Decoder architecture \citep{NIPS2014_5346} combined with Badhanau attention mechanism \citep{DBLP:journals/corr/BahdanauCB14}. The model tries to maximize the conditional probability  $P(y \mathbin{\vert} x)$ where $y$ is the target sentence and $x$ is a sequence of characters formed by the concatenation of the to-be-normalized token $w$ and context words $w_{i-k}$ to $w_{i+k}$ surrounding the token. 

The major intuition behind using two different models is that the instances of actual conversion are few and far between. Training a single deep neural network with such a heavily skewed dataset is incredibly difficult. Separating the task of predicting whether a word needs to be normalized or not and predicting the normalized sequence of words helps us increase the overall accuracy of the normalization pipeline. Both models being independent of each other are trained separately. The first model tries to maximize the classification accuracy of predicting whether a word requires normalization. The second model tries to minimize the softmax cross-entropy of the sequence of predicted words averaged over all the time-steps.

\subsection{XGBoost Classifier}

An extreme gradient boosting machine (XGBoost) model is trained to classify tokens into the following two classes: \textit{RemainSame} and \textit{ToBeNormalized}, to be used in the later stage of the pipeline. Tokenization of the training data has already been performed. Additional preprocessing of the tokens or words needs to be done before the XGBoost model can be trained on it. Specifically, we transform the individual tokens into numerical feature vectors to be fed into the model. Each token is encoded as a vector comprised of the Unicode (UTF-8) values of its individual characters. We limit the length of the vector at 30; all characters beyond this are left out. In case of shorter words, the remaining vector is filled or padded with $0$. Contextual information is incorporated into the input feature vector by prepending the preceding token and appending the succeeding token to the target input token.  To demarcate the boundaries between the consecutive tokens, we use the $-1$  integer value. Since the starting position of the target input token can vary, this value serves as the \textit{'start of token'} identifier.  An example can better elucidate the process. For instance, the input vector for \textit{'genus'} in the sentence : \textit{Brillantaisia is a genus of plant in family Acanthaceae}  will be: [ \textbf{-1,  97,}   0,   0,   0,   0,   0,   0,   0,   0,   0,   0,   0,   0,   0,   0,   0,   0,   0,   0,   0,   0,   0,   0,   0,   0,   0,   0,   0,   0,   0, \textbf{-1, 103, 101, 110, 117, 115,}   0,   0,   0,   0,   0,   0,   0,   0,   0,   0,   0,   0,   0,   0,   0,   0,   0,   0,   0,   0,   0,   0,   0,   0,   0, \textbf{-1, 111, 102,}   0,   0,   0,   0,   0,   0,   0,   0,   0,   0,   0,   0,   0,   0,   0,   0,   0,   0,   0,   0,   0,   0,   0,   0,   0,   0,   0,   0,  -1].

After the data is preprocessed and ready, we perform a train-validation split to help us tune the model. The performance metric we have used is AUC or area under the curve. A top-down approach to hyperparameter tuning is employed. We begin with a high learning rate and determine the best number of estimators or trees which is the most important hyperparameter for the model along with the learning rate. We find the best number of estimators to be 361. Then we go on to tune the tree-specific parameters such as the maximum depth of the decision stumps, the minimum child weight and the gamma value. Once we have a decent model at hand, we start tuning the regularization parameters to get better or similar performance at a reduced model complexity. Finally, we have an AUC score of 0.999875 in the training set and a score of 0.998830 on the validation set.

\begin{figure}[h]
    \centering
    \includegraphics[width=0.5\textwidth]{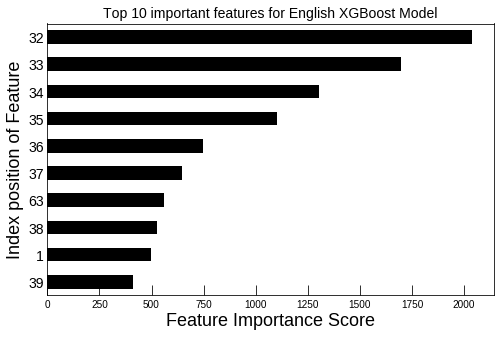}
    \caption{Top 10 important features for the English XGBoost model}
    \label{fig:feat_imp}
\end{figure}

The XGBoost package by \citep{DBLP:journals/corr/ChenG16} allows us to rank the relative importance of the features for the classification task by looking at the improvement in the accuracy brought about by any particular feature. On generating the feature importance plot of the trained English XGBoost model in Figure \ref{fig:feat_imp}, we find out that the first six characters of the target token, i.e. feature at the 32nd, 33rd, 34th, 35th, 36th and 37th position of the feature vector, had the highest score. For a human classifying these tokens, the first few characters of the word or token are indeed the best indicators to decide whether it needs to be normalized or not. This assures us that the trained model has in fact learned the right features for the task at hand. After these first few features, the model also places high importance on other characters belonging to the preceding and succeeding token. The high F1-score of the model as reported in Table \ref{tab:classification_english} confirms the overall effectiveness of the model.

The model could also be trained to classify the tokens into semiotic classes. The semiotic classes which most confuses the DNC translator could be fed to a separate sequence-to-sequence model which is exclusively trained on those error-prone classes. Another direction to go from here would be to increase the size of the context window during the data preprocessing stage to feed even more contextual information into the model. 

\subsection{Sequence to Sequence DNC}
The \textit{ToBeNormalized} tokens, as classified by the XGBoost model, are then fed to a recurrent model. For this end, we present an architecture called sequence-to-sequence DNC that allows the DNC model to be adapted for sequence-to-sequence translation purposes. Our underlying framework uses the RNN Encoder-Decoder architecture \citep{NIPS2014_5346}. We have also used attention mechanisms to allow the decoder to concentrate on the various different output states generated during the encoding phase. One major contribution of this paper is to replace bidirectional LSTM used in a Neural Machine Translation system with a single unidirectional DNC. During the encoding phase, the DNC reads an input sequence of vectors $x = (x_1,\dots, x_{T_x})$; $x_t \in \mathbb{R}^{K_x}$, and outputs a sequence of annotation vectors $h=(h_1,\dots,h_{T_x})$; $h_t \in \mathbb{R}^{n}$, where $K_x$ is the input vocabulary size and $T_x$ is the number of input tokens. 

\begin{equation}
h_t=g_e(x_t,h_{t-1},s_{t})
\end{equation}
where function $g_e$ gives the output of the DNC network during the encoding phase, and $s_{t}$ is the hidden state of the DNC. During the decoding phase the DNC is trained to generate an output word $y_t \in \mathbb{R}^{K_y}$, given a context vector $c_t \in \mathbb{R}^n$, where $K_y$ is the output vocabulary size. The decoding phase uses Bahdanau attention mechanism \citep{DBLP:journals/corr/BahdanauCB14} to generate a context vector $c_t$ by performing soft attention over the annotation vectors $h$.

The Decoder defines a conditional probability $P$ of an output word $y_t$ at time step $t$ given sequence of input vectors $x$ and previous predictions $y_{1},\dots,y_{t-1}$.

\begin{equation}\label{condprob}
P(y_t \mathbin{\vert} y_1,\dots,y_{t-1},x)=g_d(y_{t-1},s_t,c_t)
\end{equation}
where $g_d$ gives the output of the DNC network during the decoding phase, and $s_t$ is the hidden state of the of the DNC at timestep $t$ computed by
\begin{equation}
s_t=f(s_{t-1},y_{t-1},c_t)
\end{equation}
where $f$ calculates the new state of the DNC network based on the previous controller and memory states. During the decoding phase, the output of the DNC is fed into a dense layer followed by a soft-max layer to generate word-by-word predictions. We also used embedding layers to encode the input and output tokens into fixed dimensional vectors during the encoding and decoding phases. 
\begin{figure}[h]
  \centering
  
  \includegraphics[width=0.95\linewidth]{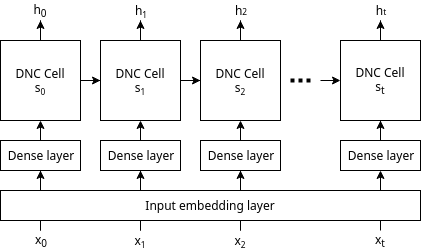}
  \caption{Sequence to sequence DNC, encoding phase.}
\end{figure}
\begin{figure}[h]
  \centering
  
  \includegraphics[width=0.95\linewidth]{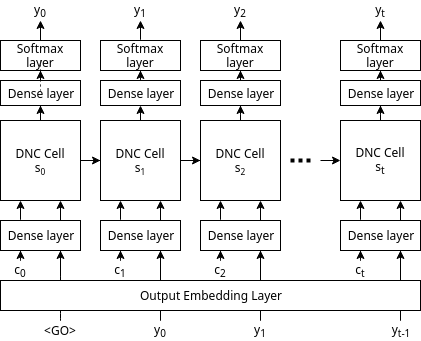}
  \caption{Sequence to sequence DNC, decoding phase.}
\end{figure}

A DNC uses a $N\times W$ dimensional memory matrix for storing state information compared to a single cell state in an LSTM. The presence of an external memory allows the DNC to store representations of the input data in its memory matrix using write heads and then read the representations from the memory using read heads. Dynamic memory allocation helps the network to encode large input sequences while retaining the inherent structure in those sequences. The content and location based attention mechanisms give the network more information about the input data during decoding. The ability to read and write from memory helps the network in meta-learning to store richer representations of the input data. We found out that network was able to generalize faster compared to LSTM with a low number of training examples. This is probably due to the fact that DNC is able to store complex quasi-regular structure embedded in the input data sequences in its memory and then later, it is able to infer from these representations.

We also found that feeding the context vector $c$ during the decoding stage was necessary for convergence. Probably, the context vector provides the network with more information about which locations to focus during each decoding step. The annotations used for generating the context vector stores information about the states of the DNC during the entire encoding phase.

For the purpose of text normalization, we feed the \textit{ToBeNormalized} tokens in a manner specified in Experiment 2 of \citep{Sproat2017RNN}. The \textit{ToBeNormalized} token is placed in between 3 context words to the left and right, with a distinctive tag marking the to-be-normalized word. This is then fed as a sequence of characters into the input embedding layer during the Encoding stage. For the sentence \textit{The city is 15kms away from here}, in order to normalize the token \textit{15km} the input becomes
\begin{center}
\texttt{
city is <norm> 15km </norm> away from
}
\end{center}

\noindent where \textit{<norm>} and \textit{</norm>} are tags that mark the beginning and end of the to-be-normalized token. The output is always a sequence of words. During decoding phase output tokens are first fed into an output embedding layer before feeding it to the decoder. For the above example, the output becomes

\begin{center}
\texttt{
fifteen kilometers
}
\end{center}

\section{Experimental Results}

The initial XGBoost classification layer gives an F1-score of 0.96 in for English and 1.00 for Russian. The classifiers' performance in terms of precision and recall of the two classes is reported in Table \ref{tab:classification_russian} and \ref{tab:classification_english}. The DNC model in the second layer was trained for 200k steps on a single GPU system with a batch-size of 16 until the perplexity reached 1.02. The parameter values used for training the second layer are given in Table \ref{tab:exp_conf} The overall results of entire system are reported in Table \ref{tab:accuracy1} \& \ref{tab:accuracy2}. It can be seen that the overall performance of the model in terms of the F1-score is quite good. The Russian model reported an accuracy of $99.3\%$ whereas the English reported an accuracy of $99.4\%$. As mentioned before the instances of actual conversion are quite a few, which is the reason for the high overall accuracy. However, when analyzing a text normalization system, it is more important to look into the kind of errors it makes. A single metric such as accuracy or BLEU (Bilingual Evaluation Understudy) score is not sufficient for comparison. It is not much of a problem if a token from the DATE class: \textit{2012} is normalized as \textit{two thousand twelve}, instead of \textit{twenty twelve}. However, we certainly would not want it to be translated to something like \textit{twenty thirteen}. These 'silly' errors are subjective by their very nature and thus, rely on a human reader. This makes the analysis of these kinds of errors difficult but important. One has to take a look at all the cases where the model produces completely unacceptable predictions.

\begin{table*}
\centering
\caption{sequence-to-sequence DNC experimental settings. $K_x$: input vocabulary size, $K_y$: output vocabulary size, $R$: number of read heads}
\begin{tabular}{lllllll} 
\toprule
               & \textbf{$K_x$} & \textbf{$K_y$} & \textbf{Memory size ($N\times W$)}  & \textbf{$R$} & \textbf{Controller hidden units} & \textbf{Embedding size} \\ 
\hline
English     & 168      & 1781   & $256\times 64$   & 5 & 1024 & 32 \\
Russian &    222   & 2578   & $256\times 64$    & 5  & 1024 & 32 \\
\bottomrule
\end{tabular}
\label{tab:exp_conf}
\end{table*}

\begin{table}[h]
\centering
\caption{Classification Report for XGBoost Russian model}
\begin{tabular}{llll} 
\toprule
               & \textbf{Precision} & \textbf{Recall} & \textbf{F1 score}  \\ 
\hline
RemainSelf     & 1.00      & 1.00   & 1.00      \\
ToBeNormalized & 0.99      & 1.00   & 1.00      \\
\bottomrule
\end{tabular}
\label{tab:classification_russian}
\end{table}

\begin{table}[h]
\centering
\caption{Classification Report for XGBoost English model}
\begin{tabular}{llll} 
\toprule
               & \textbf{Precision} & \textbf{Recall} & \textbf{F1 score}  \\ 
\hline
RemainSelf     & 1.00      & 1.00   & 1.00      \\
ToBeNormalized & 0.94      & 0.99   & 0.96      \\
\bottomrule
\end{tabular}
\label{tab:classification_english}
\end{table}

\begin{table*}
\centering 
\caption{Comparison of accuracies over the various semiotic classes of interest on the English data-set. base accuracy: accuracy of the LSTM based sequence-to-sequence model proposed in \citep{Sproat2017RNN}. accuracy: accuracy of the proposed, XGBoost + sequence-to-sequence DNC model.} 
\begin{tabular}{llrrrr}
\toprule
{} & \textbf{semiotic-class} &  \textbf{base count} &  \textbf{count} &  \textbf{base accuracy} &  \textbf{accuracy} \\
\midrule
0  &            ALL &     92416 &  92451 &          0.997 &  0.994 \\
1  &          PLAIN &     68029 &  67894 &          0.998 &  0.994 \\
2  &          PUNCT &     17726 &  17746 &          1.000 &  0.999 \\
3  &           DATE &      2808 &   2832 &          0.999 &  0.997 \\
4  &        LETTERS &      1404 &   1409 &          0.971 &  0.971 \\
5  &       CARDINAL &      1067 &   1037 &          0.989 &  0.994 \\
6  &       VERBATIM &       894 &   1001 &          0.980 &  0.994 \\
7  &        MEASURE &       142 &    142 &          0.986 &  0.971 \\
8  &        ORDINAL &       103 &    103 &          0.971 &  0.980 \\
9  &        DECIMAL &        89 &     92 &          1.000 &  0.989 \\
10 &          DIGIT &        37 &     44 &          0.865 &  0.795 \\
11 &          MONEY &        36 &     37 &          0.972 &  0.973 \\
12 &       FRACTION &        13 &     16 &          0.923 &  0.688 \\
13 &           TIME &         8 &      8 &          0.750 &  0.750 \\
\bottomrule
\end{tabular}
\label{tab:accuracy1}
\end{table*}
 
\begin{table*}
\caption{Comparison of accuracies over the various semiotic classes of interest on the Russian data-set. The headings are same as in Table \ref{tab:accuracy1}.} 
\centering
\begin{tabular}{llrrrr}
\toprule
{} & \textbf{semiotic-class} &  \textbf{base count} &  \textbf{count} &  \textbf{base accuracy} &  \textbf{accuracy} \\
\midrule
0  &            ALL &       93184 &  93196 &          0.993 &  0.993 \\
1  &          PLAIN &       60747 &  64764 &          0.999 &  0.995 \\
2  &          PUNCT &       20263 &  20264 &          1.000 &  0.999 \\
3  &           DATE &        1495 &   1495 &          0.976 &  0.973 \\
4  &        LETTERS &        1839 &   1840 &          0.991 &  0.991 \\
5  &       CARDINAL &        2387 &   2388 &          0.940 &  0.942 \\
6  &       VERBATIM &        1298 &   1344 &          1.000 &  0.999 \\
7  &        MEASURE &         409 &    411 &          0.883 &  0.898 \\
8  &        ORDINAL &         427 &    427 &          0.956 &  0.946 \\
9  &        DECIMAL &          60 &     60 &          0.867 &  0.900 \\
10 &          DIGIT &          16 &     16 &          1.000 &  1.000 \\
11 &          MONEY &          19 &     19 &          0.842 &  0.894 \\
12 &       FRACTION &          23 &     23 &          0.826 &  0.609 \\
13 &           TIME &           8 &      8 &          0.750 &  0.750 \\
\bottomrule
\end{tabular}
\label{tab:accuracy2}
\end{table*}

It can be seen that the class-wise accuracies of the model are quite similar to the base LSTM model. Upon analyzing the nature of error made in each class, it  was identified that the model performs quite well in classes: DIGIT, DATE, ORDINAL and TIME. The errors reported in these classes are shown in Table \ref{tab:errors1}. Most of the errors in these classes are due to the fact that the DNC is confused with the true context of the token. For example, the token \textit{1968} in DATE context is predicted as if in CARDINAL context. However, the DNC never makes a completely unacceptable prediction in these classes for both English and Russian data-sets as can be observed in Table \ref{tab:errors1}. For readers unfamiliar with the Russian language, we look at the issue with \foreignlanguage{russian}{22 июля}. This error stems from confusion in grammatical cases which do not exist in the English language and are replaced with prepositions. However in the Russian language prepositions are used along with grammatical cases, but may also be omitted in many situations.  Now, \textit{the 22nd} is transformed to \foreignlanguage{russian}{двадцать второго} (transliterated into Latin script as "dvadtsat vtorOGO") whereas when used with the preposition \textit{of}, such as \textit{of the 22nd},  it is transformed to  \foreignlanguage{russian}{двадцать второе}  (transliterated into Latin script as "dvadtsat vtorOE"). The baseline LSTM based sequence-to-sequence architecture proposed in Experiment 2 of \cite{Sproat2017RNN} showed completely unacceptable errors such as a DATE \textit{11/10/2008} normalized as \textit{the tenth of october two thousand eight}. On the other hand the DNC network never makes these kind of 'silly' errors in these classes. This suggests that the DNC network can, in fact, be used as a text normalization solution for these classes. This is an improvement over the baseline LSTM model in terms of the quality of prediction.

The DNC network, however, suffers in some classes: MEASURE, FRACTION, MONEY and CARDINAL, similar to the baseline LSTM network. The errors reported in these classes are shown in Table \ref{tab:errors2}. All the unacceptable mistakes in cardinals occur in large numbers greater than a million. The DNC sometimes also struggles with getting the measurement units and denominations right. For non-Russian readers, we illustrate this with the MONEY token \textit{\$1m} where the prediction is completely off; since \foreignlanguage{russian}{ одиннадцать долларов сэ ш а} means \textit{"eleven US dollars"} but \foreignlanguage{russian}{один миллион долларов сэ шэ а} means \textit{"1 million US dollars"}. In terms of overall accuracy, the DNC performs slightly better in MEASURE, MONEY and CARDINAL. The model, however, performs much worse in FRACTION compared to the baseline model. The DNC network still makes unacceptable 'silly' mistakes in these classes such as predicting completely inaccurate digits and units, which is not enough to make it a  trustworthy system for these classes. 
\begin{table*}
\caption{Errors in which the DNC network is confused with the context of the token.}
\begin{tabularx}{\textwidth}{lll>{\raggedright}X>{\raggedright}Xr}
\hline
{} & \textbf{input} & \textbf{semiotic-class} & \textbf{prediction} & \textbf{truth} & \\
\hline                              
0  &  2007 &  DIGIT & two thousand seven & two o o seven & \\
1  &  1968 & DATE & one thousand nine hundred sixty eight & nineteen sixty eight & \\
2  &  0:02:01 & TIME & zero hours two minutes and one seconds &  zero hours two minutes and one second & \\
3  & \foreignlanguage{russian}{22 июля} & DATE & \foreignlanguage{russian}{двадцать второго июля} & \foreignlanguage{russian}{двадцать второе июля} & \\
4  &  II  &  ORDINAL  &  \foreignlanguage{russian}{два} &   \foreignlanguage{russian}{второй}  &  \\
\hline
\end{tabularx}
\label{tab:errors1}
\end{table*}

\begin{table*}
\caption{Errors in which the DNC network makes completely unacceptable predictions.}
\begin{tabularx}{\textwidth}{lll>{\raggedright}X>{\raggedright}Xr}
\hline
{} & \textbf{input} & \textbf{semiotic-class} & \textbf{prediction} & \textbf{truth}  & \\
\hline                              
0  &  14356007  &  CARNINAL & one million four hundred thirty five thousand six hundred seven  & fourteen million three hundred fifty six thousand seven   &  \\
1  &   0.001251 g/cm3 & MEASURE & zero point o o one two five one sil g per hour & zero point o o one two five one grams per c c &  \\
2  &    88.5 million HRK & MONEY & eighty eight point five million yen &  eighty eight point five million croatian kunas  & \\
3  &    10/618,543 &  FRACTION & ten sixteenth sixty one thousand five hundred forty three &  ten six hundred eighteen thousand five hundred forty thirds   & \\
4 & \foreignlanguage{russian}{15 м/с}  & MEASURE & \foreignlanguage{russian}{пятнадцать сантиметров в секунду} & \foreignlanguage{russian}{ пятнадцати метров в секунду } & \\
5 & \$1m & MONEY & \foreignlanguage{russian}{ одиннадцать долларов сэ ш а} & \foreignlanguage{russian}{один миллион долларов сэ ш а} & \\
\hline
\end{tabularx}
\label{tab:errors2}
\end{table*}

In order to understand why the model performs so well in some classes but suffers in others, we proceeded to find the frequency of these specific tokens in the English training dataset. The training set has 17,712 instances of dates of the form $xx/yy/zzzz$. As reported in the earlier section, the model made zero unacceptable mistakes in these DATE tokens. The baseline LSTM, however, still reported unacceptable errors for dates of the similar form. On the other hand, measurement units such as $mA$, $g/cm3$ and $ch$ occur less than ten times in the training set. Compared to other measurement units, $kg$ and $cm$ are present more than $200$ times in the training set. CARDINAL has $273,111$ tokens out of which only 1,941 are numbers which are larger than a million. Besides, the error in MONEY for the English data-set was for the denomination that occurred only once in the training set. The results in Table \ref{tab:errors2} clearly demonstrate that the model suffers only in the tokens for which a sufficient number of examples are not available in the training set. The DNC network never made any unacceptable prediction for examples that were sufficiently present. This hints at the notion that the model performs poorly particularly for units, cardinals and denominations which occur a lesser number of times in the training set. Unlike the baseline LSTM, the model is reasonable and durable to examples which are sufficiently present.

\subsection{Ablation Study}

\begin{table*}
\caption{Class-wise accuracy comparison with and without DNC memory}
\centering
\begin{tabular}{llrr} 
\toprule
   & \textbf{semiotic-class} & \textbf{accuracy without memory} & \textbf{accuracy with memory}  \\ 
\midrule
0  & ALL            & 0.940704                & 0.994181              \\
1  & CARDINAL       & 0.345227                & 0.991321              \\
2  & DATE           & 0.186794                & 0.996469              \\
3  & DECIMAL        & 0.021739                & 1.000000              \\
4  & DIGIT          & 0.068182                & 0.818182              \\
5  & FRACTION       & 0.000000                & 0.687500              \\
6  & LETTERS        & 0.101490                & 0.971611              \\
7  & MEASURE        & 0.021127                & 0.985915              \\
8 & MONEY          & 0.054054                & 0.972973              \\
9 & ORDINAL        & 0.077670                & 0.980583              \\
10 & PLAIN          & 0.992444                & 0.993858              \\
11 & PUNCT          & 0.998873                & 0.998873              \\
12 & TIME           & 0.125000                & 0.750000              \\
13 & VERBATIM       & 0.812188                & 0.995005              \\
\bottomrule
\label{table:ablation_table}
\end{tabular}
\end{table*}

To make the case for memory augmentation in neural networks to perform text normalization, we conducted an ablation experiment to factor out its contribution if any. We know that the DNC model consists of a controller network, equipped with various memory access mechanisms to read and write from a memory matrix. During the process of training, the controller network is intended to learn to use the provided memory access mechanisms instead of just relying on its internal LSTM state. This is important to make the most out of the benefits that come from memory augmentation. The DNC controller network at each time-step receives a set of $R$ read vectors as input. These read vectors or memory activations are obtained by performing a read operation on the memory matrix. Our ablation experiment intends to verify the contributions of these memory activations for prediction. We use an existing model pre-trained in the previous section and perform inference on the test data on two conditions, with memory activations and without memory activations. The first condition is just inference on the test data without any modification to the DNC activations. For the second condition, we zero out the read vectors from the memory at each time-step before providing it as input to the controller network. If the DNC learns to use the provided memory access mechanisms during the training process (i,e, the prediction is largely dependent on the value of the read and write vectors), zeroing out the read vectors during inference should have a significant impact in the performance of the model. However, if the model is just learning to predict based on its internal LSTM controller state, zeroing out the read vectors during inference should have a minute impact in performance. As can be seen from the comparison in Table \ref{table:ablation_table}, there is a significant drop in accuracy for most semiotic classes when memory structures are inaccessible to the controller network. Particularly, semiotic classes which require structured processing of the input tokens such as, DATE, TIME, DIGIT, MEASURE, MONEY, and TELEPHONE, show a significant reduction in performance. Upon analyzing the kind of errors that the DNC network makes when memory structures are removed, it was found that the DNC network gets the translation context correct in most cases. For example, for the token \textit{1984}, the prediction of the DNC network without memory is \textit{nineteen thousand two hundred eighty one}. The model starts the translation correctly, but however, it fails mid-way by predicting a completely incorrect digit. This is indicative of our prior assumption that the model learns to write the input tokens in its memory matrix during the encoding stage and later reads from the memory during the decoding stage. If the memory structures would not have been used by the network for translation, we should have seen, essentially, no drop in performance on removing them. Apparently, the read vectors have high feature importance in performing a successful prediction during inference.
\subsection{Results on up-sampled training set}

\begin{table*}
\caption{Predictions corrected after up-sampling for the English data-set.}
\begin{tabularx}{\textwidth}{lll>{\raggedright}X>{\raggedright}Xr}
\hline
{} & \textbf{token} & \textbf{semiotic-class} & \textbf{prediction before up-sampling}                        & \textbf{prediction after up-sampling} &                  \\
\hline
0 & 295 ch         & MEASURE                 & two hundred ninety five hours                                 & two hundred ninety five chains                      &    \\
1 & 2 mA           & MEASURE                 & two a m                                                       & two milli amperes                                      & \\
2 & 0.001251g/cm3  & MEASURE                 & zero point o o one two five one sil g per hour                & zero point o o one two five one sil g per c c          & \\
3 & 1/2 cc  & MEASURE                 & one \textunderscore letter d\textunderscore letter a\textunderscore letter s\textunderscore letter h\textunderscore letter
\textunderscore letter t\textunderscore letter w\textunderscore letter o\textunderscore letter \textunderscore letter o\textunderscore letter               & one half c c         & \\
4 & 14356007       & CARDINAL                & one million four hundred thirty five thousand six hundred seven & fourteen million three hundred fifty six thousand seven & \\ 
\hline
\end{tabularx}
\label{tab:errors3}
\end{table*}   

The initial results lead us to a follow-up question. Can our system perform better given better quality data? Will a simple up-sampling procedure on the rare kinds of tokens improve the model? To test our hypothesis of whether sufficient examples can improve the performance in certain semiotic classes, we up-sampled the distribution of those specific tokens which occurred less than a particular threshold frequency. The up-sampling was done with duplication only for MEASURE and CARDINAL on the English dataset. Sentences which had measurement units which occurred less than 100 times were up-sampled to have 100 instances each in the training set. Out of 253 measurement units, 229 of them occurred less than a 100 times in the entire training set. Similarly, sentences with Cardinals with value larger than a million were up-sampled to 10,000 instances. The final distribution of the training set was 59,439 for MEASURE and 299,694 for CARDINAL. The model was then retrained for the same number of training iterations with the up-sampled data. The overall accuracies and number of unacceptable errors for MEASURE \& CARDINAL after up-sampling are shown in Table \ref{tab:accimproved}. The comparison of the predictions is shown in Table \ref{tab:errors3}. Overall, it is very interesting to see that using simple data augmentation techniques like up-sampling helped remove all the unacceptable errors in MEASURE and reduced the number of unacceptable errors from three to two in CARDINAL. Such an elementary technique even removed errors in rare measurement units such as $ch$ and $g/cc$. However, the improvement observed in CARDINAL was rather modest. And as expected, the number of unacceptable errors in other classes were unaffected. This clearly provides evidence to the initial assumption that our system improves, even if marginally, when a sufficient number of examples are produced for any particular instance type. Nonetheless, we can safely say that this system looks promising and worthy of widespread adoption.

\begin{table}
  \centering
  \caption{Accuracies and no. of unacceptable errors before and after up-sampling for the English data-set. a1: accuracy before up-sampling, a2: accuracy after up-sampling, e1: no. of unacceptable errors before up-sampling, e2: no. of unacceptable errors after up-sampling.}
  \label{tab:accimproved}
  \begin{tabular}{llllll}
  \toprule
    & \textbf{semiotic-class} & \textbf{a1} & \textbf{a2} & \textbf{e1} & \textbf{e2} \\
  \midrule
  0 & MEASURE        & 0.971                       & 0.986                     & 4                                            & 0                                           \\
  1 & CARDINAL       & 0.994                       & 0.991                     & 3                                            & 2                                           \\
  \bottomrule              
  \end{tabular}
\end{table}

\section{Discussion}
Given the reduced number of unacceptable predictions in most semiotic classes we can say that the quality of the predictions produced by DNC are much better than the baseline LSTM model. The reason why DNC works better than LSTM might be due to the presence of a memory matrix and read-write heads. The read-write heads allows the DNC to store richer representations of the data in its dynamic memory matrix. Research done by \citeauthor{Santoro2016Oneshot}, shows memory augmented neural networks have the ability to generalize well to less number of training examples. The model never made any unacceptable prediction in some classes (DIGIT, DATE, ORDINAL and TIME) for the same test set used by \citeauthor{Sproat2017RNN}. With a basic augmentation technique and minimal human requirement the number of unacceptable errors in MEASURE was reduced to zero. The LSTM model reported in \citep{Sproat2017RNN} reported unacceptable errors even when sufficient examples are present. On the other hand, DNC is quite resistant to errors when sufficient examples are present. However, the DNC is still prone to making unacceptable predictions in some classes (FRACTION, MONEY and CARDINAL) which makes it still risky as a standalone text normalization system. There is still a lot of work to be done before a purely deep learning based algorithm can be used a standalone component of a TTS system. We believe the performance the model can be further improved by designing a more balanced training set.

Apart from the domain of text normalization, we also provide evidence that a sequence-to-sequence architecture made with DNC can be successfully trained for tasks similar to machine translation systems. Until now DNC has only been used for solving simple algorithmic tasks and have not been applied to real-time production environments. The quality of the results produced by DNC in text normalization demonstrates it is, in fact, a viable alternative to LSTM based models. LSTM based architectures usually require large amounts of training data. The results in \cite{Sproat2017RNN} show that the LSTM based seq-to-seq models can sometimes produce a weird output even when sufficient examples are present. For instance, LSTM's did not work well in predicting DATE, TIME and DIGIT, even though the training set had a lot of examples from the category. DNC, on the other hand, is able to generalize well and avoids these kinds of errors. We believe that the DNC architecture should give good results in designing NMT models for languages which do not have a lot of training data available. It is also important to note that a single unidirectional DNC provides much better generalization compared to the stacked bidirectional LSTM used by \citeauthor{Sproat2017RNN}, proving that memory augmented neural networks can provide much better results with significantly reduced training times and fewer data points. The LSTM model reported in their paper was trained on 8 parallel GPUs for about five and half days (460k steps). On the contrary, our model was trained on a single GPU system for two days (200k steps). Furthermore, our model used only 2.2\% of the English data and 4.4\% of the Russian data for training.

\section{Conclusion}
Therefore, we can safely arrive at the conclusion that memory augmented neural networks such as the DNC are in fact a promising alternative to LSTM based models for a language agnostic text normalization system. Additionally, the proposed system requires significantly lesser amounts of data, training duration and compute resources. Our DNC model has reduced the number of unacceptable errors to zero for some classes with basic up-sampling of rare data points. However, there are still classes where the performance needs to be improved before an exclusively deep learning based model can become the text normalization component of a TTS system. Besides, we have also demonstrated a system that can be used to train sequence-to-sequence models using a DNC cell as the recurrent unit. 

\section*{Acknowledgements}
We would like to show our gratitude to Richard Sproat (Senior Research Scientist at Research \& Machine Intelligence, Google, New York) for his insights and comments that greatly improved the manuscript. We thank Kaggle for hosting the Text Normalization Challenge by Richard Sproat and Kyle Gorman which got us interested in this problem in the first place. We are also very grateful to Google DeepMind for open sourcing their implementation of the Differentiable Neural Computer which was a requirement for this research.

\section*{References}
\bibliographystyle{spbasic}
\bibliography{main}

\end{document}